\pdfoutput=1

\documentclass[11pt]{article}

\usepackage[final]{acl}

\usepackage{times}
\usepackage{latexsym}

\usepackage[T1]{fontenc}

\usepackage[utf8]{inputenc}

\usepackage{microtype}

\usepackage{inconsolata}

\usepackage{graphicx}
\usepackage{xspace,mfirstuc,tabulary}

\usepackage[utf8]{inputenc}
\usepackage{microtype}
\usepackage{inconsolata}
\usepackage{nicefrac}
\usepackage{float}
\usepackage[export]{adjustbox}
\usepackage[most]{tcolorbox}
\usepackage{caption}
\usepackage{array}
\usepackage{soul}
\usepackage[super]{nth}
\usepackage{mdframed}
\usepackage{inconsolata}
\usepackage{bm}
\usepackage{mathrsfs}
\usepackage{epsfig}
\usepackage{graphicx}
\usepackage{booktabs}
\usepackage{verbatim}
\usepackage{tipa}
\usepackage{algorithm}
\usepackage{url}
\usepackage{amsfonts}       
\usepackage{nicefrac}       
\usepackage{xcolor}         
\usepackage{tcolorbox}
\usepackage{algorithm}
\usepackage{listings}
\usepackage{inconsolata}
\usepackage{bm}
\usepackage{amsthm,amsmath,amssymb}
\usepackage{mathrsfs}
\usepackage{epsfig}
\usepackage{booktabs}
\usepackage{verbatim}
\usepackage{enumitem,kantlipsum}
\usepackage{lipsum}
\usepackage{tipa}
\usepackage{algorithmicx}
\usepackage{algpseudocode}
\usepackage{multirow,makecell}
\usepackage{arydshln}
\usepackage{colortbl}
\usepackage{hyperref}
\usepackage{color} 
\usepackage{dsfont}
\usepackage{tikz}
\usepackage{wrapfig}
\usepackage{xspace}
\usepackage{ifthen}
\usepackage{subcaption}
\usepackage{pdflscape}
\usepackage{pifont}

\newtcolorbox{AIbox}[2][]{aibox,title=#2,#1}

\tcbset{
  aibox/.style={
    width=450pt,
    top=10pt,
    colframe=black,
    colbacktitle=black,
    enhanced,
    center,
    attach boxed title to top left={yshift=-0.1in,xshift=0.15in},
    boxed title style={boxrule=0pt,colframe=white,},
  }
}
\definecolor{forestgreen}{rgb}{0.13, 0.55, 0.13}

\usepackage{color} 
\definecolor{mypink2}{RGB}{219, 48, 122}
\definecolor{orange}{RGB}{255, 147, 00}
\definecolor{jrcolor}{RGB}{100, 150, 225}
\definecolor{jrcomment}{RGB}{70, 200, 150}
\definecolor{grey}{RGB}{166, 166, 166}

\definecolor{mygreen}{HTML}{3cb44b}

\title{Cross-lingual Contextualized Phrase Retrieval}

\author{Huayang Li$^{\clubsuit}$\thanks{\ \ Work done during Huayang's internship at Tencent AI Lab. Correspondence to Deng Cai and Lemao Liu.}~~~~Deng Cai$^{^\heartsuit}$~~~~Zhi Qu$^{\clubsuit}$~~~~Qu Cui$^{^\heartsuit}$\\\
\textbf{Hidetaka Kamigaito}$^{\clubsuit}$~~~~\textbf{Lemao Liu}$^{^\heartsuit}$~~~~\textbf{Taro Watanabe}$^{\clubsuit}$\\
  $^\clubsuit$Nara Institute of Science and Technology~~~~
  $^\heartsuit$Tencent AI Lab~~~\\
  \texttt{\{li.huayang.lh6, qu.zhi.pv5, kamigaito.h, taro\}@is.naist.jp}\\ \texttt{\{jcykcai, qucui, redmondliu\}@tencent.com}}

\begin{document}
\maketitle
\begin{abstract}
Phrase-level dense retrieval has shown many appealing characteristics in downstream NLP tasks by leveraging the fine-grained information that phrases offer. 
In our work, we propose a new task formulation of dense retrieval,  \textit{cross-lingual contextualized phrase retrieval}, which aims to augment cross-lingual applications by addressing polysemy using context information.
However, the lack of specific training data and models are the primary challenges to achieve our goal. As a result, we extract pairs of cross-lingual phrases using word alignment information automatically induced from parallel sentences. 
Subsequently, we train our \textbf{C}ross-lingual \textbf{C}ontextualized \textbf{P}hrase \textbf{R}etriever (CCPR) using contrastive learning, which encourages the hidden representations of phrases with similar contexts and semantics to align closely.
Comprehensive experiments on both the cross-lingual phrase retrieval task and a downstream task, i.e, machine translation, demonstrate the effectiveness of \textsc{CCPR}.
On the phrase retrieval task, \textsc{CCPR} surpasses baselines by a significant margin, achieving a top-1 accuracy that is at least 13 points higher.
When utilizing \textsc{CCPR} to augment the large-language-model-based translator, it achieves average gains of 0.7 and 1.5 in BERTScore for translations from X$\Rightarrow$En and vice versa, respectively, on WMT16 dataset. 
We release our code and data at \url{https://github.com/ghrua/ccpr_release}.

\end{abstract}

\section{Introduction}

Compared with the dense retrieval at sentence (or passage) level \cite{karpukhin-etal-2020-dense,pmlr-v162-borgeaud22a, asai2024selfrag}, learning dense retrieval at the phrase level has shown more appealing characteristics in extensive NLP tasks, such as entity linking \cite{gillick-etal-2019-learning}, open-domain question answering \cite{lee-etal-2021-learning-dense,lee-etal-2021-phrase}, text generation \cite{lan2023copy, cao2024retrieval}, etc. An important reason is that phrases can provide more fine-grained information than sentences.

In cross-lingual research, the phrase-level dense retrieval also shows the promise to solve a range of NLP problems \cite{bapna-firat-2019-non, zheng-etal-2022-cross-lingual}. The cross-lingual phrase retrieval in \citet{zheng-etal-2022-cross-lingual} focuses on the mapping of cross-lingual Wiki entities, where each of them is represented by averaged hidden representations from 32 different contexts. However, the constrained phrase type, i.e., the wiki entity, makes the cross-lingual phrase retriever difficult to augment general NLP tasks. Moreover, unlike wiki entities, which may have fewer ambiguities, general-type phrases that are lexically identical can have different meanings depending on their contexts. Thus, accounting for polysemy \cite{cruse1986lexical} using the context information becomes critical.
Even when the lexically identical phrases share similar semantics, a more appropriate context would provide richer information for utilizing the phrases in downstream tasks \cite{min-etal-2019-discrete, cao2024retrieval}.

Therefore, we advocate for a new task formulation, i.e., \textit{cross-lingual contextualized phrase retrieval}, which aims to find the cross-lingual phrase that is mostly relevant to the provided (general-type) source phrases, considering their meanings and surrounding contexts. However, achieving this goal is non-trivial, due to the scarcity of specific training data and models.
Since annotating cross-lingual contextualized phrase pairs of general type is very difficult and expensive, we first introduce a data collection method, which leverages the automatically induced word alignment information from parallel sentences to extract suitable cross-lingual phrase pairs for training. This ensures that the phrases are of general type and both cross-lingual phrases and contexts are well aligned. Thereafter, we propose a \textbf{C}ross-lingual \textbf{C}ontextualized  Phrase Retriever (CCPR) which is trained on the constructed dataset. The \textsc{CCPR} mainly employs phrase-level contrastive learning to draw cross-lingual phrases with similar contexts and meanings closer in the hidden space. At inference time, we can leverage the well-trained CCPR to build a phrase-level index and use the query phases for search. 

Unlike previous works \cite{zheng-etal-2022-cross-lingual}, which focus solely on the task of cross-lingual phrase retrieval for wiki entities, our work also explores the potential of leveraging CCPR to augment downstream cross-lingual tasks, e.g., machine translation (MT). One critical question in front of us is how to select meaningful phrases for indexing and searching at inference time. To address this problem, we propose to learn a phrase segmentation module, which can be used to predict meaningful phrases from sentence- or passage-level retrieval data at inference time. The developed phrase segmentation module is important for ensuring the train-test consistency of CCPR.

Experiments on both of the cross-lingual contextualized phrase retrieval and a downstream task, i.e., MT, show the effectiveness of \textsc{CCPR}. We first evaluate \textsc{CCPR} on the cross-lingual contextualized phrase retrieval task, since a higher accuracy is more beneficial for downstream tasks, such as MT, cross-lingual dictionary induction, etc \cite{zhang2016bridging, sogaard-etal-2018-limitations}. 
For this task, we use the human annotated cross-lingual phrase pairs as the golden truth and evaluate whether \textsc{CCPR} and other baselines can retrieve those golden phrases from a large-scale index. Experiments show that \textsc{CCPR} outperforms the baselines for at least 13 points in terms of the top-1 accuracy. We also conduct evaluation on the MT task using a large language model (LLM), e.g., LLama-2 \cite{touvron2023llama}. This task aims to evaluate whether the information retrieved by \textsc{CCPR} can enhance LLM's ability of cross-lingual generation. Following the fashion of retrieval-augmented generation (RAG) \cite{lewis2020retrieval}, we simply integrate the retrieved phrase information to the input of the LLM, and compare our method with other baselines. 
For both X$\Rightarrow$En and En$\Rightarrow$X translation directions on the WMT16 dataset, where X is from six languages, our \textsc{CCPR} achieves averaged gians of 0.7 and 1.5 BERTScore points \cite{zhang2019bertscore}, respectively. Results on other evaluation metrics, e.g., COMET \cite{rei-etal-2022-comet}, are consistent with BERTScore.

In summary, our contributions are three-fold:
\begin{itemize}[noitemsep,nolistsep]
    \item We propose a new formulation of dense retrieval, i.e., \textit{the cross-lingual contextualized phrase retrieval}, which has substantial potential in augmenting cross-lingual tasks.
    \item   We propose a \textbf{C}ross-lingual \textbf{C}ontextualized \textbf{P}hrase \textbf{R}etriever (\textsc{CCPR}), which uses the constructed training data, i.e., cross-lingual phrases extracted from automatically induced word alignment information, to learn both the phrase alignment and segmentation modules.  
    \item  We conduct extensive experiments on cross-lingual contextualized phrase retrieval and MT. Our method outperforms the baselines by a large margin.
\end{itemize}

\section{Related Work}

\paragraph{Dense Retrieval} 
In text generation, dense retrieval has been widely studied at both sentence and phrase levels. Particularly after the emergence of large language models (LLMs), how to retrieve related sentences (or passages) \cite{izacard2021contriever, ni-etal-2022-sentence} for input to LLMs has became a popular research topic, namely, retrieval-augmented generation (RAG) \cite{lewis2020retrieval, karpukhin-etal-2020-dense, guu2020retrieval, pmlr-v162-borgeaud22a, asai2024selfrag}.  However, compared to the sentence-level retrieval, phrase-level retrieval has been shown to not only enhance retrieval accuracy but also markedly boost performance in downstream tasks \cite{lee-etal-2021-learning-dense,lee-etal-2021-phrase}. By indexing and integrating retrieved phrases into the output of language models, \citet{lan2023copy} and \citet{cao2024retrieval} make the text generation process more attributable and accurate. These studies underline the effectiveness of phrase-level dense retrieval. Our work notably diverges from those works, because we focus more on cross-lingual tasks.

\paragraph{Retrieval in Cross-lingual Tasks} 

In cross-lingual tasks, the majority of research has concentrated on sentence-level retrieval.  
Many research works in retrieval-augmented MT \cite{zhang-etal-2018-guiding, Gu2018SearchEG, xia2019graph, he-etal-2021-fast} have explored retrieving bilingual sentences based on similarity of source sentences, guiding MT models to utilize retrieved target information. A notable limitation is their reliance on bilingual data for constructing the index. This requirement substantially limits the scale and diversity of data available for retrieval. In contrast, \citet{cai-etal-2021-neural} propose to directly retrieve relevant target sentences using source sentences to assist MT, facilitating cross-lingual retrieval using a monolingual index. Unlike the task-specific model for MT in \citet{cai-etal-2021-neural}, many studies in dense retrieval aim to enhance the cross-lingual sentence retrieval on more languages and at a larger scale \cite{chidambaram-etal-2019-learning, conneau-etal-2020-unsupervised, reimers-gurevych-2020-making, heffernan-etal-2022-bitext, feng-etal-2022-language, cai-etal-2022-retrofitting}. 

However, cross-lingual phrase retrieval has not been extensively investigated.  \citet{bapna-firat-2019-non} utilize the source side of the bilingual data for building a retrieval index and employ the similarity between source phrases for retrieval. Therefore, their indexing also suffers from the limitation of reliance on bilingual data.  In addition, it uses $n$-grams as phrases by default, leading to plenty of meaningless and noisy spans in the index. \citet{zheng-etal-2022-cross-lingual} limit their cross-lingual phrase retrieval task to a specific phrase type, i.e., the wiki entity. The setting in their work overlooks the nuanced meanings that lexically identical phrases may convey in different contexts, making the retriever difficult to augment general NLP tasks. In contrast, our research aims to develop a cross-lingual contextualized phrase retriever for general-type phrases.

Additionally, there is research focused on employing nearest neighbor retrieval to support MT \cite{khandelwal2021nearest, zheng-etal-2021-adaptive, meng-etal-2022-fast, deguchi-etal-2023-subset}. These efforts differ from the aforementioned studies as they aim to estimate prediction distributions through nearest neighbor retrieval at each translation step.

\begin{figure*}[t]
    \begin{center}
        \includegraphics[width=1.9\columnwidth]{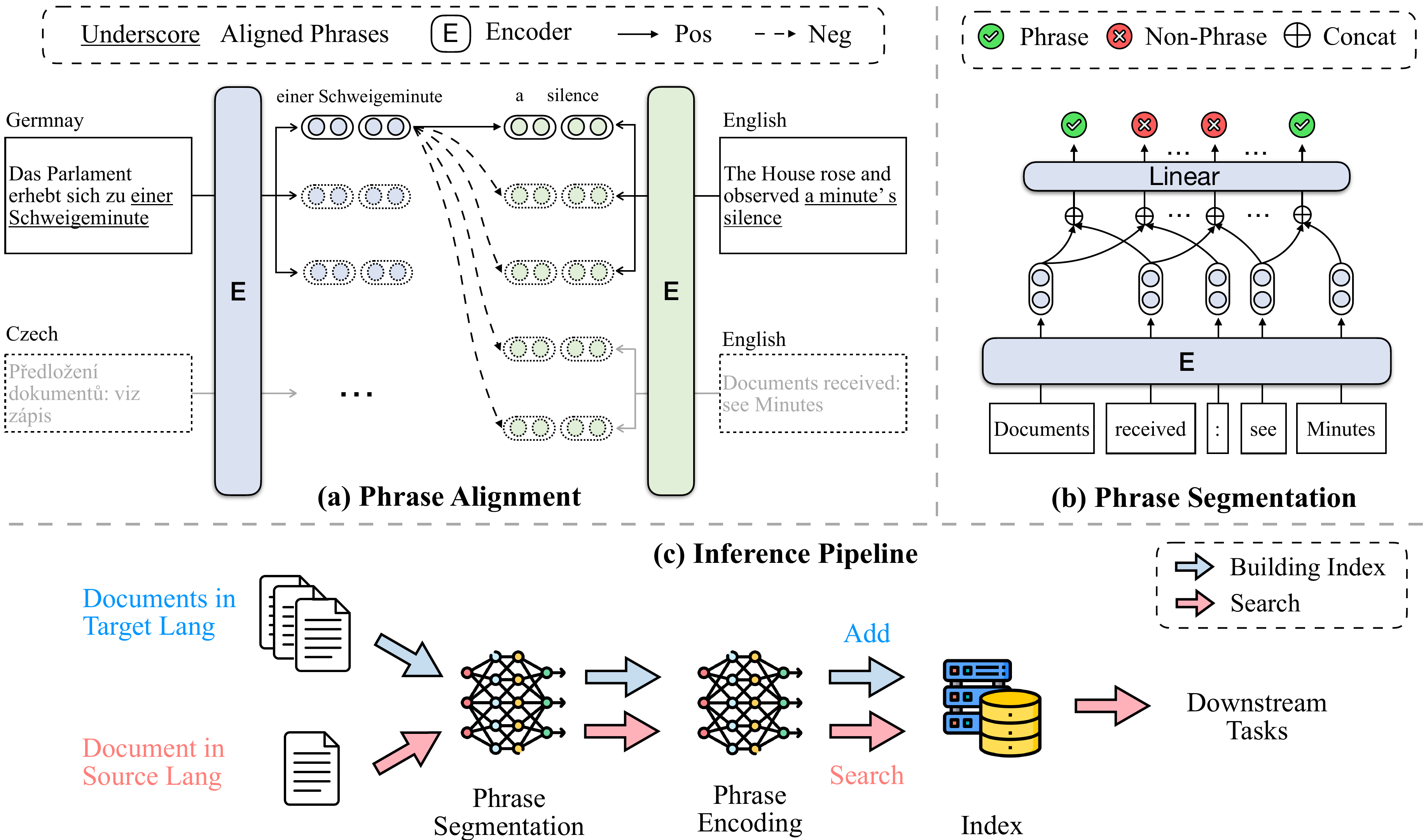}
    \end{center}
        \caption{(a) Cross-lingual phrase alignment takes the cross-lingual phrase pairs within similar contexts as positives and other in-batch phrases as negatives. Encoders with different colors employ different dropout mask $\boldsymbol{z}$, following \citet{gao-etal-2021-simcse}. The inputs of the dual encoders are parallel sentences. (b) A linear phrase-segmentation head is used to predict whether a span is a phrase or not. (c) The pipeline of using our learned model for downstream tasks. \label{fig:illustration}}
\end{figure*}

\section{Task Formulation \label{sec:task}}
This work aims a cross-lingual phrase retriever that can augment downstream NLP tasks. To ensure its effectiveness, the retriever must adhere to two essential criteria. First, it should ensure that phrases in the index are of general type. Second, it must  resolve the polysemy of general-type phrases by considering their contextual differences.

Thus, we introduce a new task formulation, \textit{cross-lingual contextualized phrase retrieval}. We are given a large collection of $N$ general-type phrases $\mathcal{P}_{index} = \{\boldsymbol{p}_1, \dots, \boldsymbol{p}_{N}\}$ in language $L_y$, and a general-type query phrase $\boldsymbol{q}$ in language $L_x$. Each phrase, by default, is associated with the context information $(\boldsymbol{c}, s, e)$, where $\boldsymbol{c}$ represents the original sentence containing this phrase, and $s$ and $e$ are its start and end positions in $\boldsymbol{c}$. The primary objective is to identify a cross-lingual phrase $\boldsymbol{p} \in \mathcal{P}_{index}$ that is relevant to $\boldsymbol{q}$, considering their contexts and meanings. Our formulation notably diverges from the one proposed by \citet{zheng-etal-2022-cross-lingual}, which focuses on a restricted type of phrases, i.e., wiki entities, ignoring the nuanced meanings that general-type phrases may convey in different contexts. 

This new task formulation also introduces several challenges, with the primary issue being the lack of training data. Specifically, there is a scarcity of cross-lingual phrase pairs of general type and accompanied by contextual information. Such data are crucial for training the model to recognize cross-lingual phrases that have similar contexts and meanings, allowing it to align these phrases closely in a hidden space. 
In addition, how to extract meaningful general-type phrases from sentences is also an open question, which is critical for employing the retriever to downstream tasks.

\section{Training Data Collection \label{sec:train_data}}

Annotating cross-lingual contextualized phrase pairs of general type is very difficult and expensive, posing a significant obstacle to training effective cross-lingual contextualized phrase retriever. However, the sentence-level parallel data in general domains are more readily available \cite{ng-etal-2019-facebook}. Additionally, as the lower unit of phrase alignment, the word alignment has been extensive studied \cite{brown-etal-1993-mathematics, och-ney-2003-systematic, dyer-etal-2013-simple, li-etal-2019-word, jalili-sabet-etal-2020-simalign, dou-neubig-2021-word, wu-etal-2023-wspalign}.  Therefore, to overcome the data scarcity, we propose to use a word alignment model, such as GIZA++ \cite{och-ney-2003-systematic} and neural aligner \cite{dou-neubig-2021-word}, in order to automatically induce word alignments from parallel sentences and subsequently extract cross-lingual phrase pairs. This method allows us to produce phrase pairs that are not only of general type but also accompanied with contexts, aligning with our task requirements.

Formally, given a pair of parallel sentences $\boldsymbol{x} = \{x_1, x_2, ..., x_{|\boldsymbol{x}|}\}$, and $\boldsymbol{y} = \{y_1, y_2, ..., y_{|\boldsymbol{y}|}\}$, where $|\boldsymbol{x}|$ and $|\boldsymbol{y}|$ are the number of words in the sequences, we use a word alignment model to obtain the word alignment information of $\boldsymbol{x}$ and $\boldsymbol{y}$ \cite{koehn-etal-2003-statistical, och-etal-1999-improved}. If every word in a consecutive span $\boldsymbol{x}_{i:j}$ can be aligned to a consecutive span $\boldsymbol{y}_{u:v}$, we will use them as a pair of aligned phrases $\boldsymbol{p}^{x}=\boldsymbol{x}_{i:j}$ and $\boldsymbol{p}^{y}=\boldsymbol{y}_{u:v}$, where $1\leq i\leq j \leq |\boldsymbol{x}|$ and $1\leq u\leq v \leq |\boldsymbol{y}|$. Notably, phrases that meet the constraints may range from a single word to the whole sequence, and extracted phrases are allowed to have overlaps. In addition, the context information of $\boldsymbol{p}^{x}$ and $\boldsymbol{p}^{y}$ are $(\boldsymbol{x}, i, j)$ and  $(\boldsymbol{y}, u, v)$, respectively. Unlike the phrase table extracted in SMT \cite{koehn-etal-2003-statistical, chiang-2005-hierarchical}, which is static and context-independent, here the phrase mapping between $\boldsymbol{p}^{x}$ and $\boldsymbol{p}^{y}$ is unique considering their context.

As shown in Fig. \ref{fig:illustration} (a), The Germany phrase (``einer'', ``Schweigeminute'') is aligned with the the English phrase (``a'', ``minute'', ``'s'', ``silence'') within similar contexts, i.e., parallel sentences. More details about the extraction of cross-lingual contextualized phrase pairs are in App.  \ref{app:phrase_extraction}.

\section{Methodology }

\subsection{Model Architecture \label{sec:alignment}}
We introduce a \textbf{C}ross-lingual \textbf{C}ontextualized \textbf{P}hrase \textbf{R}etriever (CCPR) for our new formulated task. The main target of CCPR is to make the representations of cross-lingual phrases with similar contexts and meanings to be close in the hidden space. To this end, we propose a \textit{cross-lingual phrase alignment} module based on contrastive learning, utilizing the data collected in Sec.  \ref{sec:train_data} for training. In addition, one remaining problem is how to select phrases for indexing at inference time. To address this, we propose a \textit{phrase segmentation} module that uses phrase representations to predict meaningful phrases from sentences or paragraphs. At inference time, we can leverage the learned phrase segmentation module to select phrases for indexing, ensuring the train-test consistency.

\paragraph{Phrase Encoding}
Our phrase encoder is based on a context encoder and an $\mathrm{MLP}$ layer. 
The encoding of a phrase $\boldsymbol{p}=\boldsymbol{x}_{s:e}$, is defined as:
\begin{align}
    \mathbf{H}^{z} &= \mathrm{ContextEncoder}(\boldsymbol{x}, \boldsymbol{z}) \nonumber \\
    \boldsymbol{h}^{z}_{p} &= \mathrm{MLP}_{align}\big([\mathbf{H}^z_{s}; \mathbf{H}^z_{e}]\big) \label{eq:enc}
\end{align}
where $\mathrm{ContextEncoder}(\boldsymbol{x}, \boldsymbol{z})$ is a Transformer model, e.g., BERT \cite{devlin-etal-2019-bert, feng-etal-2022-language} and RoBERTa \cite{conneau-etal-2020-unsupervised}, that encodes the context $\boldsymbol{x}$ to a matrix $\mathbf{H}^{z} \in \mathbb{R}^{N\times d}$ with a dropout mask $\boldsymbol{z}$. Inspired by preview works \cite{lan2023copy, lee-etal-2021-learning-dense, seo-etal-2018-phrase}, we use the concatenation of $\mathbf{H}^{z}_{s}$ and $\mathbf{H}^{z}_{e}$, i.e., the start and end token of the phrase, to represent phrase $\boldsymbol{x}_{s:e}$.
$\mathrm{MLP}_{align}(\cdot)$ is a function that maps the concatenation of two hidden states $\mathbf{H}^{z}_{s}$ and $\mathbf{H}^{z}_{e}$ from $\mathbb{R}^{2d}$ to $\mathbb{R}^{o}$, where $o$ is the output hidden size. The encoding for phrases in context $\boldsymbol{y}$ is similar.

\paragraph{Cross-lingual Phrase Alignment} For a batch of parallel sentences, we can extract a collection of cross-lingual phrase pairs $\mathcal{P}_{pair} = \{(\boldsymbol{p}^{x}_{i}, \boldsymbol{p}^{y}_{i})\}_{i=1}^{K}$, where $\boldsymbol{p}^{x}_{i}$ and $\boldsymbol{p}^{y}_{i}$ are the positive examples of each other. Since each positive phrase pair is from parallel sentences that share similar semantics, directly learning on them may cause the model to learn a trivial shortcut. Therefore, inspired by SimCSE \cite{gao-etal-2021-simcse}, we apply two independently sampled dropout mask $\boldsymbol{z}$ and $\boldsymbol{z}^\prime$, to encode phrases $\boldsymbol{p}^{x}_{i}$ and $\boldsymbol{p}^{y}_{i}$, respectively, following Eq. \ref{eq:enc}. The training objective for the $x\rightarrow y$ direction is:

\vspace{-0.5em}
\begin{align}
    \mathcal{L}_{x\rightarrow y} &= -\frac{1}{K}\sum_{i=1}^K\log\frac{\exp(\boldsymbol{h}^{z}_{p^{x}_i}\cdot \boldsymbol{h}^{z'}_{p^{y}_i})}{Z_x(i)}, \\
   Z_x(i) &= \sum_{(p^x_j, p^y_j) \in \mathcal{P}_{pair}}\exp(\boldsymbol{h}^{z}_{p^{x}_i}\cdot \boldsymbol{h}^{z}_{p^{y}_j}) \label{eq:con_deno} 
\end{align}
where we use all the non-paired phrases in $\mathcal{P}_{pair}$ as in-batch negatives in Eq. \ref{eq:con_deno}. For bidirectional symmetry, our final loss is:

\begin{equation}
    \mathcal{L}_{align} = \mathcal{L}_{x\rightarrow y} + \mathcal{L}_{y\rightarrow x}.
\end{equation}

The illustration of the contrastive learning is shown in Fig. \ref{fig:illustration} (a). In practice, we may use parallel data from multiple languages during training, i.e., the source contexts $\boldsymbol{x}$ in a batch may come from multiple languages, e.g., Germany, Czech, etc, which is inspired by the success of multi-lingual training in \citet{liu-etal-2020-multilingual-denoising}.

\paragraph{Phrase Segmentation\label{sec:seg}}
This module aims to learn how to select phrases from sentences or passages, since most retrieval data at inference time comes in sentences or passages, rather than phrases.
To ensure the train-test consistency, we take the phrases extracted in Sec.  \ref{sec:train_data} as positive data, and all the other spans in the corresponding contexts as negative data. Our phrase segmentation module is a binary classifier defined as follows:
\begin{align}
    P(T=1) = \sigma\Big(\mathrm{MLP}_{seg}\big([\mathbf{H}^z_{i}; \mathbf{H}^z_{j}]\big)\Big)\label{eq:pseg} 
\end{align}
where $T$ is the label for the span $\boldsymbol{x}_{i:j}$, where $1\leq i\leq j \leq |\boldsymbol{x}|$, $\sigma(\cdot)$ is an activation function, $\mathrm{MLP}_{seg}$ is linear layer that maps the hidden representation from $\mathbb{R}^{2d}$ to $\mathbb{R}$, and the definitions for $\mathbf{H}^z_{i}$ and $\mathbf{H}^z_{j}$ are the same as in Eq. \ref{eq:enc}. We use the label $T=1$ for phrases and $T=0$ for non-phrase spans. The phrase segmentation for sequence $\boldsymbol{y}$ is similar to Eq. \ref{eq:pseg}. In practice, the number of non-phrase spans is significantly more than the number of phrases. To mitigate data imbalance, we employ a strategy of randomly sampling an equal number of non-phrase spans and phrases within a sentence during training \cite{li2020empirical}. The illustration of phrase segmentation is in Fig. \ref{fig:illustration} (b). 

The training loss for the phrase segmentation is
\vspace{-0.5em}
\begin{align}
    \mathcal{L}_{seg} = -\frac{1}{|\mathcal{S}|}&\sum_{\boldsymbol{p} \in \mathcal{S}} \Big( T_{p}\log P(T_{p}=1) + \\
    & (1-T_p)\log\big(1-P(T_{p}=1)\big) \Big) \nonumber
\end{align}
where $\mathcal{S}$ is a set of spans, which are extracted from sequence $\boldsymbol{x}$ or $\boldsymbol{y}$. The learning objective of the whole model becomes:

\begin{equation}
    \mathcal{L} =  \mathcal{L}_{align} + \beta \mathcal{L}_{seg} 
\end{equation}
where $\beta$ is a hyper-parameter. It is worth noting that the parameters of $\mathrm{ContextEncoder}$ used for phrase alignment and segmentation are shared.

In our preliminary studies, we investigated various phrase segmentation strategies, including $n$-gram segmentation and learning a Byte Pair Encoding (BPE) model \cite{sennrich-etal-2016-neural, kudo-2018-subword} across word boundaries. However, these methods  struggle to identify meaningful phrases, leading to significant discrepancies between training and testing phases and performance declines.
In Sec. \ref{sec:mt}, we evaluate the effect of the learned phrase segmentation on MT task.

\begin{figure}[t]
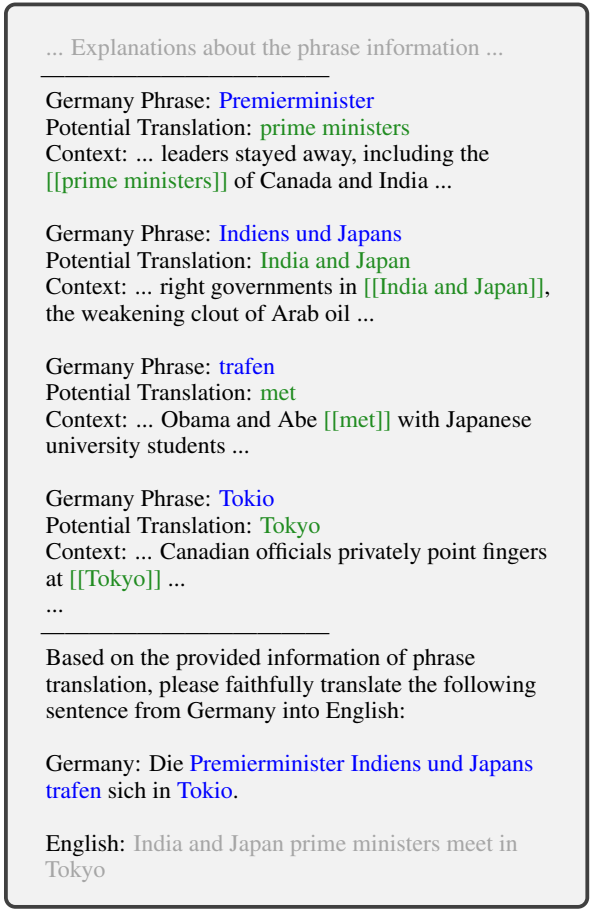
\centering
\tiny
\begin{minipage}{1.0\columnwidth}\vspace{0mm}    \centering
\begin{tcolorbox} 
    \raggedright
    \small
     \hspace{-6mm}
    \  \\
    \textcolor{grey}{... Explanations about the phrase information ...} \\
    ------------------------------------ \\
Germany Phrase: {\color{blue}Premierminister}\\
Potential Translation: {\color{forestgreen} prime ministers} \\
Context: ... leaders stayed away, including the {\color{forestgreen}[[prime ministers]]} of Canada and India ... \\
\ \\
Germany Phrase: {\color{blue}Indiens und Japans} \\
Potential Translation: {\color{forestgreen}India and Japan} \\
Context: ... right governments in {\color{forestgreen}[[India and Japan]]}, the weakening clout of Arab oil ... \\
\ \\
Germany Phrase:  {\color{blue}trafen} \\
Potential Translation:  {\color{forestgreen}met} \\
Context: ... Obama and Abe {\color{forestgreen}[[met]]} with Japanese university students ... \\
\ \\
Germany Phrase:  {\color{blue}Tokio} \\
Potential Translation:  {\color{forestgreen}Tokyo} \\
Context: ... Canadian officials privately point fingers at {\color{forestgreen}[[Tokyo]]} ... \\
... \\
------------------------------------ \\
    Based on the provided information of phrase translation, please faithfully translate the following sentence from Germany into English: \\
    \ \\
    Germany: Die {\color{blue}Premierminister} {\color{blue}Indiens und Japans} {\color{blue}trafen} sich in {\color{blue}Tokio}. \\ 
\ \\
 English: \textcolor{grey}{India and Japan prime ministers meet in Tokyo}

\end{tcolorbox}
    
\caption{An example instruction for a large-language-model based translator augmented by our method. The segmented phrases in the source sentence are in \textcolor{blue}{blue text}. The retrieved translation of the phrases are in \textcolor{forestgreen}{green}, and their appearances in the contexts are marked by ``\textcolor{forestgreen}{[[]]}''. The \textcolor{grey}{text}  after ``English:'' is the reference, which is for illustration and will not appear in the instruction. \label{tab:ours_prompt}}
\end{minipage}
\end{figure}

\subsection{Inference Pipeline\label{sec:inference}}

The pipeline of employing our cross-lingual contextualized phrase retriever for downstream NLP tasks involves two main steps: index building and searching, as shown in Fig. \ref{fig:illustration} (c).

A significant benefit of our retriever is its ability to leverage monolingual data in the target language, eliminating the need for bilingual data to construct the index at inference time \cite{zhang-etal-2018-guiding, Gu2018SearchEG, bapna-firat-2019-non}. Given the retrieval data in target language, our retriever begins by segmenting  sentences (or passages) into phrases. When segmenting a sentence (or passage), we will enumerate all the possible spans within it and calculate their probabilities of being  meaningful phrases, according to Eq. \ref{eq:pseg}. We select all the spans whose probabilities are larger than a threshold for indexing. It is worth noting that overlapping phrases are allowed at the phrase segmentation time. The selected phrases are encoded into hidden representations according to Eq. \ref{eq:enc}, which will be used to construct a dense-retrieval index.

For querying, we first process the query sentence (or paragraph) in a source language similarly, i.e., segmenting and encoding. Subsequently, we use the encoded query phrases to directly search for the related cross-lingual phrases using Maximum Inner Product Search (MIPS).  Because of the advanced data structure and search algorithm \cite{malkov2018efficient, douze2024faiss, johnson-2022-binary}, the search step is highly efficient (more details in App. \ref{app:inf_latency}). The retrieved information, including the target phrase, its accompanied context, and its positions in the context, can then be integrated to downstream tasks to improve the performance. As shown in Fig. \ref{tab:ours_prompt},  the retrieved phrases along with their surrounding contexts are integrated into the instructions given to an LLM for tasks such as MT.

\begin{table}[]
\centering
\large
\resizebox{1.0\columnwidth}{!}{
\begin{tabular}{l|ccc|c}
\toprule
\textbf{Model} & \textbf{De}$\Rightarrow$\textbf{En} & \textbf{Ro}$\Rightarrow$\textbf{En} & \textbf{Cs}$\Rightarrow$\textbf{En} & \textbf{AVG.}\\\hline
\textsc{XLMR}   & 1.5 & 0.0 & 4.0 & 1.8 \\
\textsc{mBERT} & 8.0 & 3.0 & 7.5 & 6.1 \\
\textsc{mUSE} & 34.5 & 30.0 & 45.5 & 36.6 \\
\textsc{LaBSE}  & 52.0 & 33.5 & 61.0 & 48.8 \\\cdashline{1-5}
\textsc{CPR-XLMR}  & 54.0 & 63.0 & 73.0 & 63.3 \\
\textsc{CPR-LaBSE} & 57.0 & 60.0 & 78.5  &  65.1 \\\cdashline{1-5}
\textsc{CCPR-XLMR}  & 74.5 & \textbf{73.5} & 82.5 & 76.8 \\
\textsc{CCPR-LaBSE} & \textbf{75.0} & 72.5 & \textbf{88.0} &  \textbf{78.5} \\\bottomrule
\end{tabular}}
\caption{Cross-lingual contextualized phrase retrieval. We use the accuracy@1 as our evaluation metric. The \textsc{CCPR} and \textsc{CPR} denote contextualized and context-independent cross-lingual phrase retriever, respectively. The Best results are highlighted in \textbf{bold text}. \label{tab:retrival}}
\end{table}

\section{Experiments}
We propose to evaluate our model on cross-lingual contextualized phrase retrieval and MT.

\subsection{Implementation Details \label{sec:impl}}

\paragraph{Training Data Collection}

We use all the bilingual training datasets of WMT16 on Huggingface\footnote{\url{https://huggingface.co/datasets/wmt16}} to train our \textbf{C}ross-lingual \textbf{C}ontextualized \textbf{P}hrase \textbf{R}etriever (CCPR). The WMT16 dataset has six language pairs, inclduing De-En, Cs-En, Fi-En, Ru-En, and Tr-En. Because of the efficiency and the satisfactory performance of GIZA++ \cite{och-ney-2003-systematic, dou-neubig-2021-word}, we first use the GIZA++ software\footnote{\url{https://github.com/moses-smt/mgiza}}, i.e., IBM-4 model \cite{brown-etal-1993-mathematics}, to induce the word alignment for each pair of parallel sentences, and then extract the cross-lingual phrase pairs.  More details about phrase extraction are in App. \ref{app:phrase_extraction}.
The final training dataset contains 1.3 billion cross-lingual phrase pairs extracted from 10 million parallel sentences.

\paragraph{Model}
In our work, we train two variants of our  model, \textsc{CCPR-XLMR} and \textsc{CCPR-LaBSE}, whose parameters are initialized from  XLM-RoBERTa-base (\textsc{XLMR}) \cite{conneau-etal-2020-unsupervised} and \textsc{LaBSE} \cite{feng-etal-2022-language}, respectively. The hidden size $d$ of both models is 768. The output $o$ of $\mathrm{MLP}_{align}: \mathbf{R}^{2d} \rightarrow \mathbf{R}^{o}$ in Eq. (\ref{eq:enc}) is 128, which plays a critical role in reducing the memory cost of our phrase index while maintaining comparable performance.
At inference time, we use the FAISS library\footnote{\url{https://github.com/facebookresearch/faiss}} \cite{douze2024faiss} to build our retrieval index. We use the \texttt{FlatIP} as our index type. Training details are App. in \ref{app:train_details}

\begin{table*}[]
\centering
\tiny
\resizebox{2.0\columnwidth}{!}{
\begin{tabular}{l|cccccc|c}
\toprule
 \textbf{Model} (X$\Rightarrow$En) & \textbf{De}$\Rightarrow$\textbf{En} & \textbf{Cs}$\Rightarrow$\textbf{En} & \textbf{Fi}$\Rightarrow$\textbf{En} & \textbf{Ru}$\Rightarrow$\textbf{En} & \textbf{Ro}$\Rightarrow$\textbf{En} & \textbf{Tr}$\Rightarrow$\textbf{En} & \textbf{AVG.} \\\hline
\textsc{Llama-2-7B} & 75.42 & 69.03 & 66.13 & 71.65 & 72.89 &    55.10 & 68.37 \\
\ \ + \textsc{XLMR} & 75.32 & 69.15 & 66.06 & 71.74 & 73.07 & 55.49 & 68.47 \\
\ \ + \textsc{LaBSE} & 75.74 & 69.60  & 66.81 & 72.12 & 73.22 &   56.76 & 69.04 \\\cdashline{1-8}
\ \ + \textsc{CCPR-XLMR} & 75.84 & \textbf{70.11} & 66.70 & 71.49 & 73.53 &  58.44 & 69.35 \\
\ \ + \textsc{CCPR-LaBSE} & \textbf{75.96} & 70.05 & \textbf{67.17} & \textbf{72.28} & \textbf{73.57} & \textbf{59.22} & \textbf{69.70} \\\hline\hline
\textbf{Model} (En$\Rightarrow$X) & \textbf{En}$\Rightarrow$\textbf{De} & \textbf{En}$\Rightarrow$\textbf{Cs} & \textbf{En}$\Rightarrow$\textbf{Fi} & \textbf{En}$\Rightarrow$\textbf{Ru} & \textbf{En}$\Rightarrow$\textbf{Ro} & \textbf{En}$\Rightarrow$\textbf{Tr} & \textbf{AVG.} \\\hline
\textsc{Llama-2-7B} & 63.42 & 52.32 & 48.96 & 84.18 & 82.99 & 58.47 & 65.05  \\
\ \ + \textsc{XLMR} & 64.03 & 52.34 & 49.04 & 84.08 & 82.99 & 58.66 & 65.19 \\
\ \ + \textsc{LaBSE} & 64.26 & 52.90 & 49.76 & 84.28 & 83.05 & 60.02 & 65.71 \\\cdashline{1-8}
\ \ + \textsc{CCPR-XLMR} &  \textbf{64.42} & 54.53 & 51.12 & 84.93 & 83.18 &  63.40 & 66.92\\
\ \ + \textsc{CCPR-LaBSE} & 64.29 & \textbf{54.93} & \textbf{51.54} & \textbf{84.97} & \textbf{83.36} & \textbf{64.07} & \textbf{67.19} \\
\bottomrule
\end{tabular}}\caption{MT on the test sets of WMT16. For all the retrieval-based method, the index is built on the monolingual newscrawl data of the target language. We use the BERTScore \cite{zhang2019bertscore} as the evaluation metric. The best performance of each translation direction is highlighted in \textbf{bold text}. Results on additional evaluation metrics, e.g., COMET \cite{rei-etal-2020-comet}, are in App. \ref{app:mt} \label{tab:mt}}
\end{table*}

\subsection{Cross-lingual Contextualized Phrase Retrieval \label{sec:ccpr_exp}}
As outlined in Sec.  \ref{sec:task}, given a query phrase and its context in source language, the objective of this task is to identify the most relevant cross-lingual phrase from a large-scale index in target language, considering the context information.
\paragraph{Setup} We build the test set based on the human annotated word alignment data \cite{jalili-sabet-etal-2020-simalign}. We process the word alignment data on three language pairs, De$\Rightarrow$En \cite{ghader-monz-2017-attention}, Ro$\Rightarrow$En \cite{mihalcea-pedersen-2003-evaluation}, and Cs$\Rightarrow$En \cite{marevcek2011automatic}. For the Cs$\Rightarrow$En dataset, we only leverage the data under the ``pcedt'' split, which are data from The Wall Street Journal (WSJ). For each language pair, we instruct human annotators to identify 200 high-quality phrase pairs. The details of the human annotation process are discussed in App. \ref{app:human}. These annotated source phrases serve as queries, while their aligned target phrases help construct the index. To mimic real-world conditions, where an index contains extra data,  we use 9.6 million English phrases sampled from the WMT16 training dataset as the extra data. 

We compare our models, \textsc{CCPR-XLMR} and \textsc{CCPR-LaBSE}, with several baselines. The first two baselines \textsc{XLMR} \cite{conneau-etal-2020-unsupervised}  \& \textsc{mBERT} \cite{devlin-etal-2019-bert} \footnote{\url{https://huggingface.co/models}} are models trained on multilingual data. We use them to encode the context and represent the phrases by concatenating the hidden representations of their start and end tokens. The second two baselines \textsc{mUSE} \cite{chidambaram-etal-2019-learning} \& \textsc{LaBSE} \cite{feng-etal-2022-language} are trained on cross-lingual data, and we use the model weights released by SBERT\footnote{\url{https://www.sbert.net}} \cite{reimers-gurevych-2020-making} to encode the contexts. To evaluate the effect of context-awareness, we propose two context-independent counterparts, i.e., \textsc{CPR-XLMR} and \textsc{CPR-LaBSE}. Compared with our contextualized models, the only difference is that the \textsc{CPR-X} methods use cross-lingual phrase pairs in semantically different contexts as positive examples, similar to the setup in \citet{zheng-etal-2022-cross-lingual}.
We build an independent index for each model. In this task, we use the top-1 accuracy to evaluate our methods and baselines.

\begin{table}[]
\centering
\resizebox{1\columnwidth}{!}{
\begin{tabular}{lc|ccc|l}
\toprule
\textbf{Model} & \textbf{Index} &  \textbf{En}$\Rightarrow$\textbf{Ro} & \textbf{En}$\Rightarrow$\textbf{Cs}  & \textbf{En}$\Rightarrow$\textbf{Tr} & \textbf{AVG.} \\\hline
\textsc{CCPR-XLMR} & Train & 82.89 & 53.66 & 62.03 & 66.19 \\
\textsc{CCPR-XLMR} & NC &  \textbf{83.18} & \textbf{54.53} & \textbf{63.40} & \textbf{67.03} \\\cdashline{1-6}
\textsc{CCPR-LaBSE} & Train & 83.19 & 54.23 & 63.22 & 66.88 \\
\textsc{CCPR-LaBSE} & NC & \textbf{83.36} & \textbf{54.93} & \textbf{64.07} & \textbf{67.45} \\\bottomrule
\end{tabular}}
\caption{Analysis about the retrieval data for indexing. The ``Train''  and ``NC'' indicate using the target sides of WMT16 training data and monolingual newscrawl data to build the index, respectively. We use BERTScore as our evaluation metric.\label{tab:monolingual}}
\end{table}

\paragraph{Results} Tab. \ref{tab:retrival} shows that our models significantly outperform baselines in cross-lingual contextualized phrase retrieval, highlighting the capability of our models to accurately identify relevant cross-lingual phrases while being context-aware.  In contrast, the \textsc{CPR-LaBSE} and \textsc{CPR-XLMR}, i.e., the context-independent baselines inspired by \citet{zheng-etal-2022-cross-lingual}, perform notably worse.

\subsection{Machine Translation\label{sec:mt}}

The evaluation on MT task is to assess if the information retrieved by our models can enhance performance on downstream tasks. Rather than training a new MT model \cite{vaswani2017attention}, we propose to integrate the information to the prompt of the LLM, e.g., Llama-2 \cite{touvron2023llama}, following the fashion of retrieval-augmented generation (RAG) \cite{lewis-2020-rag}. Thus, we can fairly evaluate the effect of the retrieved information.

\paragraph{Setup} 
We choose \textsc{Llama-2-7B} \cite{touvron2023llama} model as the backbone model for MT. We use two of our phrase retrieval models for this task, i.e., \textsc{CCPR-Labse} and \textsc{CCPR-XLMR}, as discussed in Sec. \ref{sec:impl}. We also consider a multilingual and a cross-lingual sentence retrieval models as baseline methods, i.e., the XLM-RoBERTa-base (\textsc{XLMR}) \cite{conneau-etal-2020-unsupervised} and \textsc{LaBSE} \cite{feng-etal-2022-language}. The two sentence-level baselines are proposed to evaluate whether the phrase-level retrieval information is more beneficial for augmenting LLM's cross-lingual ability.

We integrate the retrieval information into input of the LLM. An example of the prompt of our method is shown in Fig. \ref{tab:ours_prompt}, where the retrieved target phrases with their contexts are presented along with the source phrases in the prompt. The presented contexts are truncated to no more than 100 characters in our method.  More details about prompts of our method and other baselines are shown in App. \ref{app:prompt}. For our methods, when building the index, we first use the learned segmentation module (Sec. \ref{sec:seg}) to extract contextualized phrases from the retrieval data and then encode them into dense vectors using our models. For the \textsc{XLMR} and \textsc{LaBSE} baselines, we directly use them to encode sentences in retrieval data and build a sentence-level index.
For all methods, we use the monolingual newscrawl data of 2016 \footnote{\url{https://data.statmt.org/news-crawl/${LANG}/news.2016.${LANG}.shuffled.deduped.gz}} as the retrieval data. More setup details are in App. \ref{app:mt_all}.

\paragraph{Main Results} 

As shown in Tab. \ref{tab:mt}, both of our \textsc{CCPR-XLMR} and \textsc{CCPR-LaBSE} outperform the baseline methods when assisting the LLM with the information from cross-lingual contextualized phrase retrieval. Especially on some low-resource directions, such as the Tr$\Leftrightarrow$En, En$\Rightarrow$Cs, En$\Rightarrow$Fi, \textsc{CCPR-LaBSE} outperforms other baselines by more or around 2 BERTScore points. This indicates that the the contextualized cross-lingual phrase information is more critical for improving the performance on low resource languages, because the percentage of Cs, Fi, and Tr data in Llama-2 training are all less than $0.03\%$ \cite{touvron2023llama}. Consistent results on COMET \cite{rei-etal-2022-comet} are shown in App. \ref{app:mt}.

In addition, we find a good correlation between the performance of our cross-lingual contextualized phrase retrieval task and the MT task, showing that higher retrieval accuracy leads to superior results in downstream applications.

\begin{table}[]
\centering
\resizebox{1.0\columnwidth}{!}{
\begin{tabular}{l|ccc|c}\toprule
\textbf{Segmentation} & \textbf{En}$\Rightarrow$\textbf{Ro}             &  \textbf{En}$\Rightarrow$\textbf{Cs}                   & \textbf{En}$\Rightarrow$\textbf{Tr}     & \textbf{AVG.}              \\\midrule
N-gram & 83.16  & 53.42 & 60.74 & 65.77 \\
\textsc{PSM}  &  \textbf{ 83.36} & \textbf{54.93} & \textbf{64.07} & \textbf{67.45}
\\\bottomrule\end{tabular}
    }
\caption{Analysis of phrase segmentation methods. The $N$-gram method uses all available 5-grams for indexing and searching, whereas \textsc{PSM} employs the learned phrase segmentation module. The extracted phrases are encoded by the same encoder, i.e., \textsc{CCPR-LaBSE}. We use BERTScore as the evaluation metric.\label{tab:tok}}
\end{table}

\paragraph{Analysis} In Tab. \ref{tab:mt}, all retrieval-based methods construct their index using the monolingual newscrawl data, which is approximately six times the size of the bilingual training data.  It highlights a key benefit of our cross-lingual phrase retrieval: the ability to utilize extensive monolingual resources to build the index \cite{cai-etal-2021-neural}. Therefore, in Tab. \ref{tab:monolingual}, we evaluate the differences of indexing on monolingual data and the bilingual data. We observe that exploring the vast monolingual data leads to significantly better performance. In addition, although the $N$-gram is widely used for phrase segmentation in previous works \cite{bapna-firat-2019-non, lee-etal-2021-learning-dense, lee-etal-2021-phrase, lan2023copy}, our experiments demonstrate that a learned phase segmentation is more suitable for augmenting cross-lingual tasks, as shown in Tab. \ref{tab:tok}.

\section{Conclusion \& Future Works}
This paper introduces a novel approach to dense retrieval, i.e., cross-lingual contextualized phrase retrieval, focusing on resolving phrase polysemy by utilizing contextual information. The main challenge identified is the scarcity of training data, specifically cross-lingual phrase pairs with context. To overcome this, we use a word alignment model to derive such phrase pairs from parallel sentences. We then present the Cross-lingual Contextualized Phrase Retriever (CCPR), which employs contrastive learning to effectively capture similar meanings and contexts of cross-lingual phrases. Our extensive testing across retrieval and machine translation tasks shows that CCPR significantly outperforms existing baselines.

\section{Limitations}
While our cross-lingual contextualized phrase retrieval holds substantial potential, it is not without limitations. Notably, the phrase-level index required by our approach is considerably larger than that of a sentence-level index, given that each sentence can encompass numerous phrases. This expansion necessitates increased disk space for storing the index, requiring additional engineering techniques to maintain the scalability of our cross-lingual contextualized phrase retriever (CCPR), e.g., index quantization or sharding. Furthermore, to improve the performance of large language models (LLMs) on cross-lingual tasks, it becomes necessary to integrate the information from multiple retrieved phrases of the query sentence into the LLM input. This integration process can lead to a rise in the inference costs associated with LLMs.

\section{Ethical Statement}
We focus on leveraging cross-lingual contextualized phrase retrieval to augment the performance of downstream NLP tasks. We emphasize that our model is strictly designed and applied in a manner that avoids the generation of sensitive information, such as disinformation or content aimed at deceiving individuals. Furthermore, we assure that all data utilized for the training and evaluation of our model have been sourced from publicly accessible datasets. Our commitment to ethical research ensures our work benefits the NLP field responsibly, without compromising the ethical standards.

\bibliography{custom, anthology}

\clearpage
\appendix

\begin{table*}[t]
\centering
\tiny
\resizebox{1.8\columnwidth}{!}{
\begin{tabular}{l|llllll|l}
\toprule
Model & De-En & Cs-En & Fi-En & Ru-En & Ro-En & Tr-En & AVG. \\\midrule
\textsc{Llama-2-7B} & 86.019  & 82.509 & 84.955 & 83.616 & 85.005 & 77.551 & 83.27 \\
\ \ + \textsc{XLMR} & 85.915 & 82.571 & 84.889 & 83.575 & 84.984 & 77.809 &  83.29 \\
\ \ + \textsc{LaBSE} & 86.079 & 82.840  & 85.136 & 83.750 & 85.059 &  78.552 &  83.56 \\\cdashline{1-8}
\ \ + \textsc{CCPR-XLMR} & \textbf{86.217} & \textbf{83.079} & 85.231 & 83.745 & 85.285 & 79.502 & 83.84 \\
\ \ + \textsc{CCPR-LaBSE} & 86.202 & 83.014 & \textbf{85.342} & \textbf{83.855} &\textbf{85.322} & \textbf{80.123} & \textbf{83.97} \\\midrule\midrule
 & En-De & En-Cs & En-Fi & En-Ru & En-Ro & En-Tr \\\midrule
\textsc{Llama-2-7B} & 82.396  &  79.131 & 82.516 &  84.191 & 82.588  &  58.959 & 78.29 \\
\ \ + \textsc{XLMR} & 82.821 & 78.888 & 82.442 & 84.097 & 82.802 &  59.095 & 78.35 \\
\ \ + \textsc{LaBSE} & 83.128 & 79.533 & 83.214 & 84.403 & 82.849  &  61.183 & 79.05 \\\cdashline{1-8}
\ \ + \textsc{CCPR-XLMR}  & \textbf{83.529} & 80.517 & 83.962 & \textbf{85.377} & 83.776 & 65.975 & 80.52 \\
\ \ + \textsc{CCPR-LaBSE}  & 83.042 & \textbf{80.860} & \textbf{84.095} & 85.308 &  \textbf{84.031} & \textbf{67.060} & \textbf{80.73}
\\\bottomrule
\end{tabular}}\caption{Machine Translation. COMET.\label{tab:comet}}
\end{table*}

\begin{table}[H]\centering
\tiny
\begin{minipage}{1.0\columnwidth}\vspace{0mm}    \centering
\begin{tcolorbox} 
    \raggedright
    \small
     \hspace{-6mm}
    \  \\
    \textcolor{grey}{... Explanations about the retrieved sentence ...} \\
    ------------------------------------ \\
Related English Sentence: The Prime Ministers of India and Pakistan recently met in Pakistan to discuss the question. \\
... \\
------------------------------------ \\
    Based on the provided related sentence, please faithfully translate the following sentence from Germany into English, and do not alter its meaning: \\
    \ \\
    Germany: Die Premierminister Indiens und Japans trafen sich in Tokio. \\ 
\ \\
 English: \textcolor{grey}{India and Japan prime ministers meet in Tokyo}

\end{tcolorbox}
    
\caption{An exemplar instruction of the LLM augmented by cross-lingual sentence retrieval. The \textcolor{grey}{text}  after ``English:'' is the reference, which will not appear in the instruction. \label{tab:sent_prompt}}
\end{minipage}
\end{table}

\begin{table}[H]\centering
\tiny
\begin{minipage}{1.0\columnwidth}\vspace{0mm}    \centering
\begin{tcolorbox} 
    \raggedright
    \small
     \hspace{-6mm}\\
 Please faithfully translate the following sentence from Germany into English, and do not alter its meaning: \\
    Germany: Die Premierminister Indiens und Japans trafen sich in Tokio. \\ 
 English: \textcolor{grey}{India and Japan prime ministers meet in Tokyo}

\end{tcolorbox}
    
\caption{An exemplar instruction of the vanilla LLM. The \textcolor{grey}{text}  after ``English:'' is the reference, which is just for illustration and will not appear in the instruction. \label{tab:llm_prompt}}
\end{minipage}
\end{table}

\section{Implementation Details}
\subsection{Extraction of Cross-lingual Phrase Pairs\label{app:phrase_extraction}}
Because of the efficiency and the satisfactory performance of GIZA++ \cite{och-ney-2003-systematic, dou-neubig-2021-word}, we first use the GIZA++ software\footnote{\url{https://github.com/moses-smt/mgiza}}, i.e., IBM-4 model \cite{brown-etal-1993-mathematics}, to induce the word alignment for each pair of parallel sentences, and then extract the cross-lingual phrase pairs.
Notably, we did not induce the phrase table as in Phrase-based Statistical Machine Translation (PBSMT) \cite{koehn-etal-2003-statistical}. Instead, in our setting, we extract the cross-lingual consecutive spans that are aligned in a pair of parallel sentence as cross-lingual phrase pairs. Each of the cross-lingual phrase pairs is associated with the surrounding contexts. We filter out the phrases which begin or end with words whose frequency is more than 30k in the training dataset. We also filter out the phrases that contain only numbers and punctuations.

\subsection{Training of CCPR\label{app:train_details}}
Each model is trained on the mixture of all language pairs of WMT16. We train our model on 8 V100 GPUs for 20K steps, where the learning rate is 5e-5, dropout rate is 0.2, batch size on each device is 64. The $\beta$ for the phrase segmentation loss in Eq. \ref{eq:pseg} is $1$. Those hyper-parameters are tuned on the validation dataset.

\subsection{Inference Latency\label{app:inf_latency}}
In our experiments of cross-lingual contextualized phrase retrieval, retrieving the top-32 nearest neighbors for 1000 source phrases took only 0.04 seconds on 4 V100 GPUs, thanks to FAISS's high parallelism on GPUs \cite{douze2024faiss}. We believe this retrieval latency is adequate for most real-time retrieval scenarios.

\section{Cross-lingual Contextualized Phrase Retrieval}

\subsection{Human Annotation\label{app:human}}
We asked human annotators to label 200 high-quality phrase pairs as the golden truth data for each of the language pairs, i.e., De$\Rightarrow$En, Ro$\Rightarrow$En, and Cs$\Rightarrow$En. Hiring human annotators to annotate those bilingual data is expensive and time consuming. However, fortunately, some bilingual datasets already provide the human-annotated word alignment  \cite{ghader-monz-2017-attention, mihalcea-pedersen-2003-evaluation, marevcek2011automatic}. Therefore, in our task, we only ask the human annotators, three authors of our work, to annotate the English side of the data, i.e., whether a English span is a high-quality and meaningful phrase or not. 

More concretely, we first use heuristic rules as discussed in Sec. \ref{sec:impl} to collect an initial set of cross-lingual phrase pairs. For each phrase pair, we ask the human annotator to answer three questions:
\begin{enumerate}
    \item If the phrase is a single word, whether it is informative in the context?
    \item If the phrase has multiple words, whether the semantics of this phrase is complete and informative? For instance, ``local authorities and large'' is not a phrase with complete semantics. In addition, ``Of course'' has complete semantics but is not informative.
\end{enumerate}
If the answer of any questions is true, then we add this phrase to a pool of high-quality phrases. For those phrases, overlaps are allowed. We finally randomly sampled 200 annotated phrases form the pool for each language.

\section{Machine Translation\label{app:mt_all}}

\subsection{Translation Instructions\label{app:prompt}}
The instructions for the vanilla large language model (LLM), e.g., \textsc{ Llama-2}, and the LLM augmented by cross-lingual sentence retrieval are shown in Tab. \ref{tab:sent_prompt} and \ref{tab:llm_prompt}, respectively. 

\subsection{Building Index\label{app:mt_index}}

\begin{table}[]
\resizebox{1.0\columnwidth}{!}{
\begin{tabular}{l|ccc}\toprule
                     & \textbf{En$\Rightarrow$De} & \textbf{En$\Rightarrow$Cs} & \textbf{En$\Rightarrow$Ro} \\\midrule
CoT & 55.20               & 47.02               & 80.22               \\\midrule
CCPR-LaBSE    & 64.29               & 54.93               & 83.36 \\\bottomrule
\end{tabular}}
\caption{Translation results based on \textsc{Llama-7-7B}. For the CoT method, we use the template in Figure \ref{tab:cot_prompt}, and all the rest setups are the same as other methods in our work. We use the BERTScore as the metric \label{tab:cot_result}}
\end{table}

The indexing strategies diverge between the baseline methods and our approach. For sentence-level baselines, i.e., \textsc{XLMR} and \textsc{LaBSE}, the indexing is straightforward, directly using the sentences. In contrast, our method employs a learned phrase segmentation module to extract phrases from sentences for indexing. When applying our model to MT, we set the phrase segmentation thresholds in Eq. \ref{eq:pseg} to $0.7$ for indexing and $0.9$ for querying. The rationale behind a lower indexing threshold is to populate the index with a broader array of phrases, enhancing its comprehensiveness. Conversely, a higher threshold for query documents ensures that the segmented phrases are more accurate. These hyper-parameters were tuned using the validation dataset from WMT16.

\subsection{Format Alignment \label{app:format_align}}
Since the output format of the Llama-base model is hard to control and  the Llama-chat model, i.e., the model after instruction-tuning, may reject to translate a sentence that contains negative words, we decide to fine-tune the Llama base model using LoRA \cite{hu2022lora} for 100 steps to align the translation format. We use Platypus\footnote{\url{https://github.com/arielnlee/Platypus}} \cite{platypus2023} as our code base. We follow all the hyper-parameters of Platypus to fine-tune the LoRA module. The data we used for fine-tuning contain 1,800 parallel sentences randomly sampled from the WMT16 training set. For each method, we only fine-tune one model for the X$\Rightarrow$En (or En$\Rightarrow$X) direction, where X is the non-English language.

\subsection{More Results\label{app:mt}}

More results on evaluated by COMET \cite{rei-etal-2020-comet} are shown in Tab. \ref{tab:comet}, which is consistent with the results evaluated by BERTScore \cite{zhang2019bertscore}.

Additional, to evaluate that the effectiveness of the retrieved information, we also compare with a Chain-of-Though (COT) approach \cite{wei2022chain}. As shown in Figure \ref{tab:cot_result}, simply translating phrases before sentence translation does not achieve the expected performance gain and is slightly worse than the baseline approach. In contrast, our approach outperforms this method by a large margin. One hypothesis is that our approach is able to leverage external knowledge, which is more helpful for the translation task.

\begin{table}\centering
\tiny
\begin{minipage}{1.0\columnwidth}\vspace{0mm}    \centering
\begin{tcolorbox} 
    \raggedright
    \small
     \hspace{-6mm}\\
 \#\#\# Instruction:\\
Please faithfully translate the following sentence from English into Germany, and do not alter its meaning.\\
\ \\
English: \\
Sumptuous accommodations, the ultimate in Guest Services and a relaxing, professional atmosphere are beckoning at The Ambiance. \\
\ \\
\#\#\# Output: \\
Let's translate the sentence step by step. Firstly, we can translate the English phrases as follows:\\
\ \\
English Phrase: Sumptuous accommodations\\
Germany Phrase: prächtige Unterkünfte\\
\ \\
(more phrases)\\
...\\
Based on the phrase-level translation, the sentence-level Germany translation is: \\
\ \\
\textcolor{grey}{Kostspielige, prächtige Unterkünfte, Serviceleistungen höchsten Standards und eine entspannende Atmosphäre erwarten Sie im The Ambiance.}\\
\end{tcolorbox}
    
\caption{An exemplar template for the translation with Chain-of-Thought. The \textcolor{grey}{text} is the reference, which is just for illustration and will not appear in the input. \label{tab:cot_prompt}}
\end{minipage}
\end{table}

\end{document}